\documentclass[10pt,twocolumn,letterpaper]{article}

\usepackage{cvpr}
\usepackage{times}
\usepackage{epsfig}
\usepackage{graphicx}
\usepackage{amsmath}
\usepackage{amssymb}

\usepackage{subfig}
\usepackage{xcolor}
\usepackage{tabularx}

\usepackage[breaklinks=true,bookmarks=false]{hyperref}

\cvprfinalcopy 

\def\cvprPaperID{****} 

\begin{document}
  
  \title{PydMobileNet: Improved Version of MobileNets with Pyramid Depthwise Separable Convolution}
  
  \author{Van-Thanh Hoang\\
    Department of Electrical Engineering,\\
    University of Ulsan\\
    Ulsan, South Korea\\
    {\tt\small thanhhv@islab.ulsan.ac.kr}
    \and
    Kang-Hyun Jo\\
    Department of Electrical Engineering,\\
    University of Ulsan\\
    Ulsan, South Korea\\
    {\tt\small acejo@ulsan.ac.kr}
  }
  
  \maketitle
  
  \begin{abstract}
    Convolutional neural networks (CNNs) have shown remarkable performance in various computer vision tasks in recent years. However, the increasing model size has raised challenges in adopting them in real-time applications as well as mobile and embedded vision applications. Many works try to build networks as small as possible while still have acceptable performance. The state-of-the-art architecture is MobileNets. They use Depthwise Separable Convolution (DWConvolution) in place of standard Convolution to reduce the size of networks. This paper describes an improved version of MobileNet, called Pyramid Mobile Network. Instead of using just a $3\times 3$ kernel size for DWConvolution like in MobileNet, the proposed network uses a pyramid kernel size to capture more spatial information. The proposed architecture is evaluated on two highly competitive object recognition benchmark datasets (CIFAR-10, CIFAR-100). The experiments demonstrate that the proposed network achieves better performance compared with MobileNet as well as other state-of-the-art networks. Additionally, it is more flexible in fine-tuning the trade-off between accuracy, latency and model size than MobileNets.
  \end{abstract}
  
  \section{Introduction}
  
  Deep convolutional neural networks (CNNs) have shown remarkable performance in many computer vision tasks in recent years. The primary trend for solving major tasks is building deeper and larger CNNs \cite{girshick2014rich,he2016deep,szegedy2015going}. The most accurate CNNs usually have hundreds of layers and thousands of channels \cite{he2016deep,huang2017densely,szegedy2016rethinking,zagoruyko2016wide}. Many real-world applications need to be performed in real-time and/or on limited-resource mobile devices. Thereby, the model should be compact and low computational cost. The model compression work is actually investigating the trade-off between efficiency and accuracy.
  
  Recently, many research work focus on the field of model compression \cite{howard2017mobilenets,iandola2016squeezenet,rastegari2016xnor,wu2016quantized,zhang2018shufflenet}. These works can be separated into two main kinds of approaches: compressing existing architecture with pre-trained models and designing new efficient architectures that will be trained from scratch. The compressing approach usually bases on traditional compression techniques such
  as hashing \cite{chen2015compressing}, Huffman coding \cite{han2015deep}, factorization \cite{jaderberg2014speeding}, pruning \cite{see2016compression}, and product quantization \cite{wu2016quantized}.
  
  The second approach actually has already been investigated earlier than the first one. Inspired by the architecture proposed in \cite{lin2013network}, the Inception module is proposed in GoogLeNet \cite{szegedy2015going} to build deeper networks without increase model size and computational cost. Then it is further improved in \cite{szegedy2016rethinking} through factorizing convolution. The Depthwise Separable Convolution (DWConvolution) generalized the factorization idea and decomposed the standard Convolution into a depthwise convolution followed by a pointwise $1\times 1$ convolution. MobileNets \cite{howard2017mobilenets,sandler2018mobilenetv2} and other networks \cite{chollet2017xception,zhang2018shufflenet} have designed CNNs for mobile devices based on DWConvolution and shown that this operation to be able to achieve comparable results with fewer parameters.
  
  This paper focuses on the second approach and proposes an improved version of MobileNets \cite{howard2017mobilenets} and MobileNetV2 \cite{sandler2018mobilenetv2}, called Pyramid MobileNets (PydMobileNet), by using a pyramid kernel size for DWConvolution instead of just a $3\times 3$ kernel size to capture more spatial information. The bottleneck-liked architecture of Residual block \cite{he2016identity} is used to control \#channels of DWConvolution. Additionally, there are two ways to combine the output of pyramid DWConvolution which are addition and concatenation. Therefore, the proposed network can be from very thin to very thick. It means there are many efficient ways to investigate the trade-off between accuracy, latency, and model size for PydMobileNets.
  
  \section{Related Work and Background}
  \begin{table*}[!t]
    \def\arraystretch{1.1}%
    \setlength\tabcolsep{10pt}
    \caption{Structure of Networks for benchmarking with CIFAR-10 and CIFAR-100 datasets. The Residual block can be with standard Convolution, Depthwise Separable Convolution, Addition or Concatenation Pyramid Depthwise Separable Convolution. The first residual block of all stage (excepts stage 1) has stride = 2, others have stride = 1. The output of the Classifier layer can be 10 or 100, corresponding to dataset CIFAR-10 or CIFAR-100.}
    \label{tbl:structure}
    \centering
    \begin{tabular}{l||c|c|c}
      \hline
      Group & Output size & Net-29 & Net-56\\\hline
      \hline
      Image & $32\times 32 \times 3$ & \multicolumn{2}{c}{} \\\hline
      Convolution & $32\times 32 \times 32$ & \multicolumn{2}{c}{$3\times 3$ convolution} \\\hline
      Stage 1 & $32\times 32 \times 32$ & Residual block $\times$ 3 & Residual block $\times$ 6\\\hline
      Stage 2 & $16\times 16 \times 64$ & Residual block $\times$ 3 & Residual block $\times$ 6\\\hline
      Stage 3 & $8\times 8 \times 128$ & Residual block $\times$ 3 & Residual block $\times$ 6\\\hline
      Pooling & $1\times 1 \times 128$ & \multicolumn{2}{c}{$8\times 8$ Global Average Pooling}\\\hline
      Classifier & $1\times 1 \times 10/100$ & \multicolumn{2}{c}{10/100D fully-connected}\\\hline
    \end{tabular}
  \end{table*}
  
  \subsection{Related Work}
  This section briefly introduces about two main approaches of model compression: compressing existing architecture and designing an efficient architecture.
  
  \paragraph{Compressing existing architecture.} Most of works applied this approach improves the inference efficiency of CNNs via weight quantization \cite{hubara2016binarized,rastegari2016xnor} and/or weight pruning \cite{hassibi1993optimal,he2017channel,lecun1990optimal}. This approach is effectual because the deep CNNs usually have a substantial number of redundant weights which can be quantized or pruned without reducing (and sometimes can be even improving) accuracy. Different pruning or quantizing techniques may lead to different levels of granularity \cite{mao2017exploring}. The coarse-grained pruning methods such as filter-level pruning \cite{alvarez2016learning,he2017channel} have not a high degree of sparsity, but the output networks are much more regular, which facilitates efficient implementations and can be run in any kind of devices. In contrast, the fine-grained pruning, e.g., independent weight pruning \cite{han2015learning,lecun1990optimal}, generally achieves a higher degree of sparsity. However, it requires storing a large number of indices and also relies on special hardware/software accelerators, means hard to be implemented in real applications.
  
  \paragraph{Designing efficient architectures.} Recently, there are many studies focus on this approach \cite{howard2017mobilenets,huang2017densely,iandola2016squeezenet,sandler2018mobilenetv2,zhang2017interleaved,zhang2018shufflenet,zoph2017learning}. They have explored efficient CNNs that can be trained end-to-end. Three well-known applicants of this kind of approach that are sufficiently efficient to be deployed on mobile devices are MobileNet \cite{howard2017mobilenets,sandler2018mobilenetv2}, ShuffleNet \cite{sandler2018mobilenetv2}, and Neural Architecture Search networks (NASNet) \cite{zoph2017learning}. All these networks use DWConvolutions, which greatly reduce computational requirements without significantly reducing accuracy. A practical downside of these networks is DWConvolution are not (yet) efficiently implemented in most prominent deep-learning platforms. Therefore, some studies use the well-supported group convolution operation [25], such as CondenseNet \cite{huang2018condensenet} and Res-NeXt \cite{xie2017aggregated}, leading to better computational efficiency in practice.
  
  Besides these two main approaches, there is another approach, called \textbf{architecture-agnostic efficient inference}, which does not compress model actually, but try to reduce the inference time. The prominent examples of this approach are knowledge distillation \cite{buciluǎ2006model,hinton2014distilling,radosavovic2018data} and dynamic inference methods \cite{bolukbasi2017adaptive,figurnov2017spatially,huang2017multi}. The knowledge distillation methods train small ``student" networks to reproduce the output of large ``teacher" networks to reduce inference-time costs. And dynamic inference methods adapt the inference to each specific test example, skipping units or even entire layers to reduce computation. These methods do not be explored here but they can be used in the proposed network as well as any methods belong to the two main approaches.
  
  \begin{table}[!t]
    \def\arraystretch{1.1}%
    \setlength\tabcolsep{10pt}
    \caption{Residual block transforming from $d_i$ to $d_j$ channels, with stride $s$ and width multiplier $\alpha$.}
    \label{tbl:residualblocktransform}
    \centering
    \begin{tabular}{c|c|c}
      Input & Operator & Output \\\hline
      \hline
      $h\times w\times d_i$ & $1\times 1$ conv2d & $h\times w\times \alpha d_i$ \\
      $h\times w\times \alpha d_i$ & Conv2d/DWConv(s) & $\dfrac{h}{s}\times \dfrac{w}{s}\times \alpha d_i$ \\
      $\dfrac{h}{s}\times \dfrac{w}{s}\times \alpha d_i$ & $1\times 1$ conv2d & $\dfrac{h}{s}\times \dfrac{w}{s}\times d_j$ \\
    \end{tabular}
  \end{table}
  
  \begin{figure*}[!t]
    \begin{minipage}{\linewidth}
      \centerline{
        \includegraphics[width=0.3\linewidth]{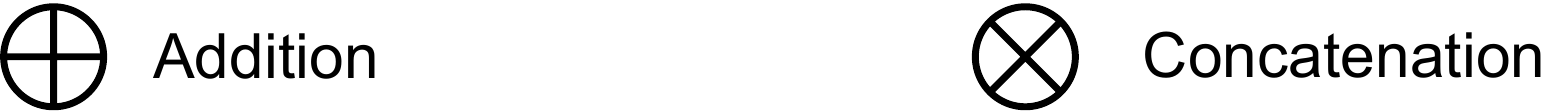}
        \label{fig:legend}}
      \centering
      \subfloat[Residual block with standard Convolution]{\includegraphics[width=0.19\linewidth]{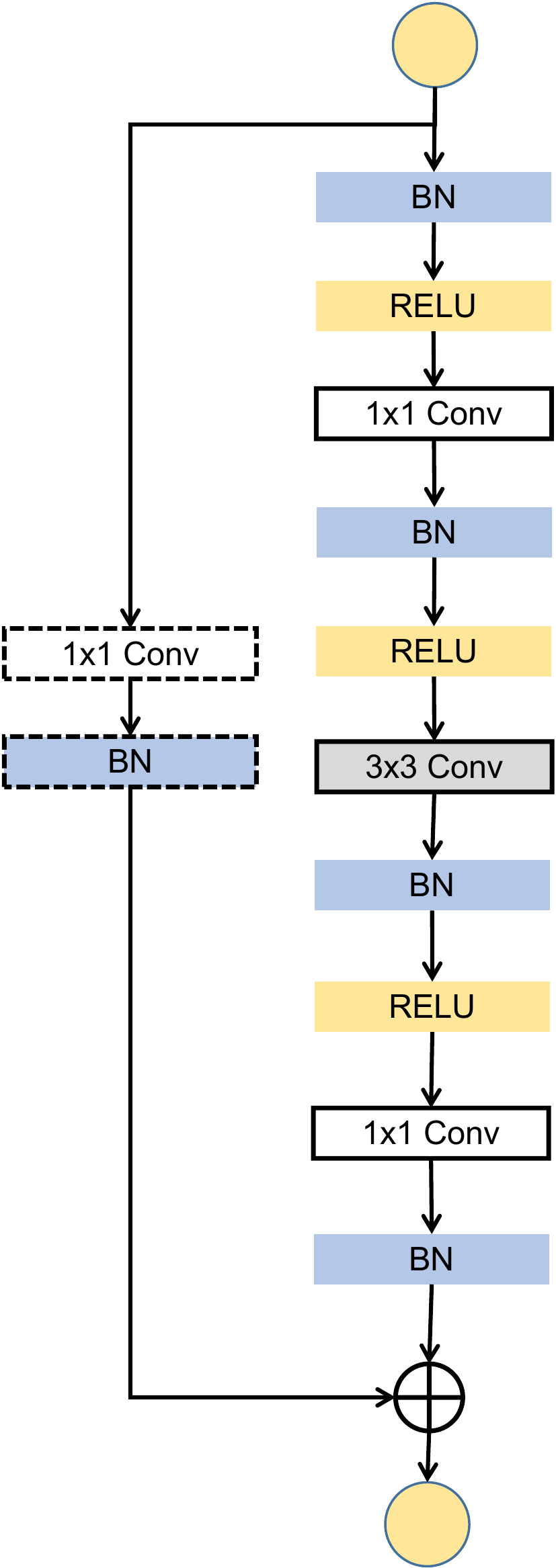}
        \label{fig:residual}}
      \hspace{0.02\linewidth}
      \subfloat[Residual block with Depthwise Convolution]{\includegraphics[width=0.19\linewidth]{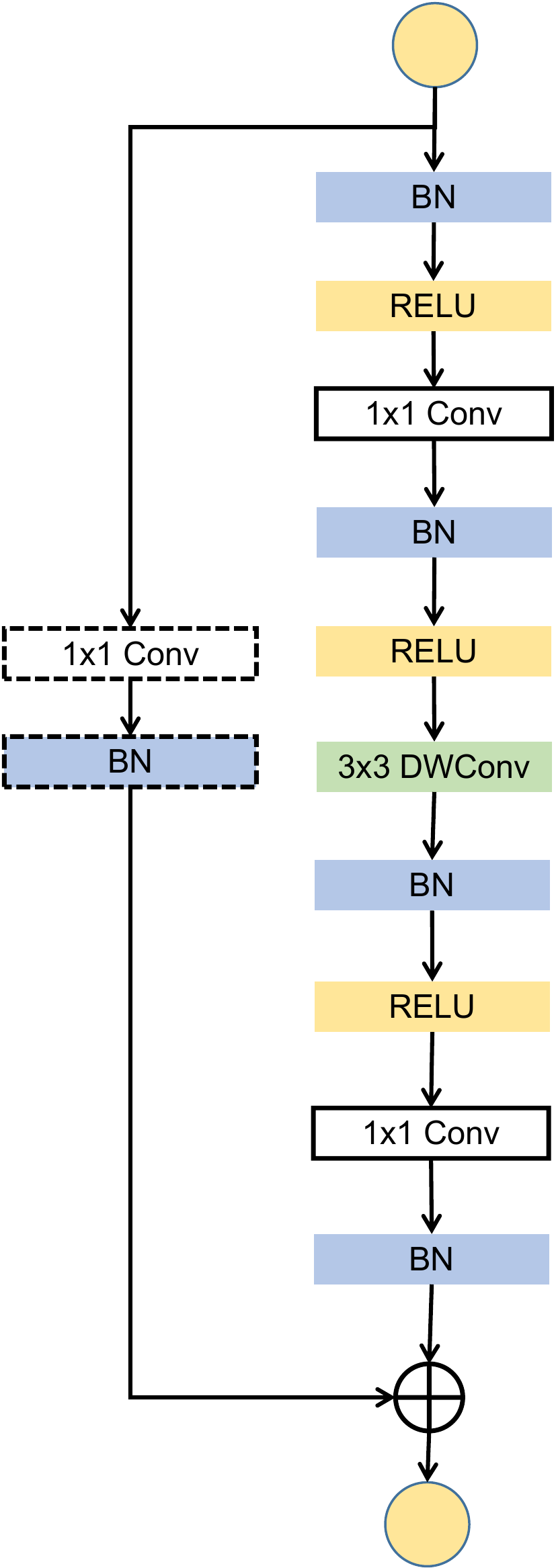}
        \label{fig:mobile}}
      \hspace{0.02\linewidth}
      \subfloat[Residual block with Addition Pyramid Depthwise Convolution ]{\includegraphics[width=0.265\linewidth]{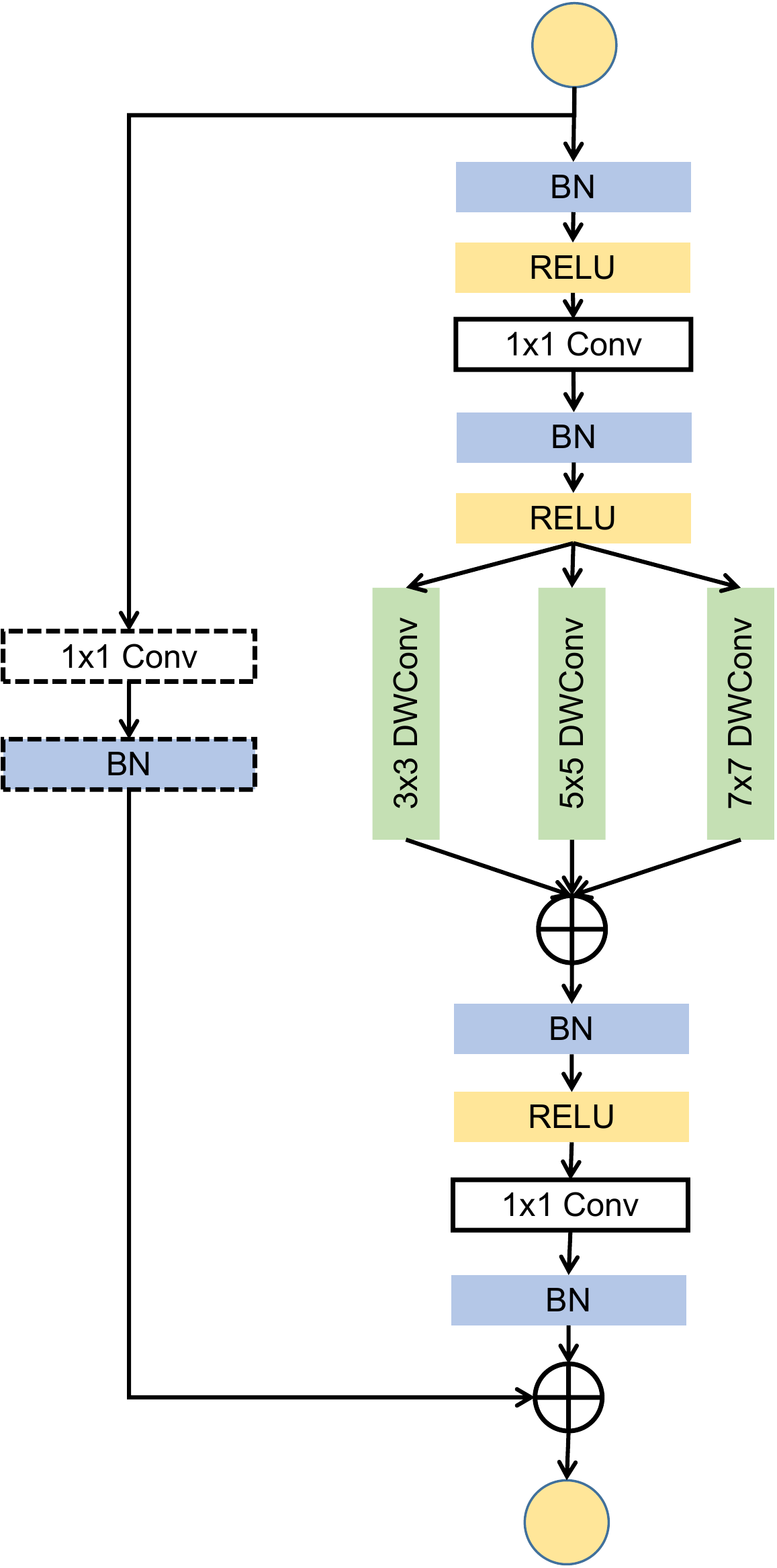}
        \label{fig:msmobile_plus}}
      \hspace{0.01\linewidth}
      \subfloat[Residual block with Concatenation Pyramid Depthwise Convolution ]{\includegraphics[width=0.265\linewidth]{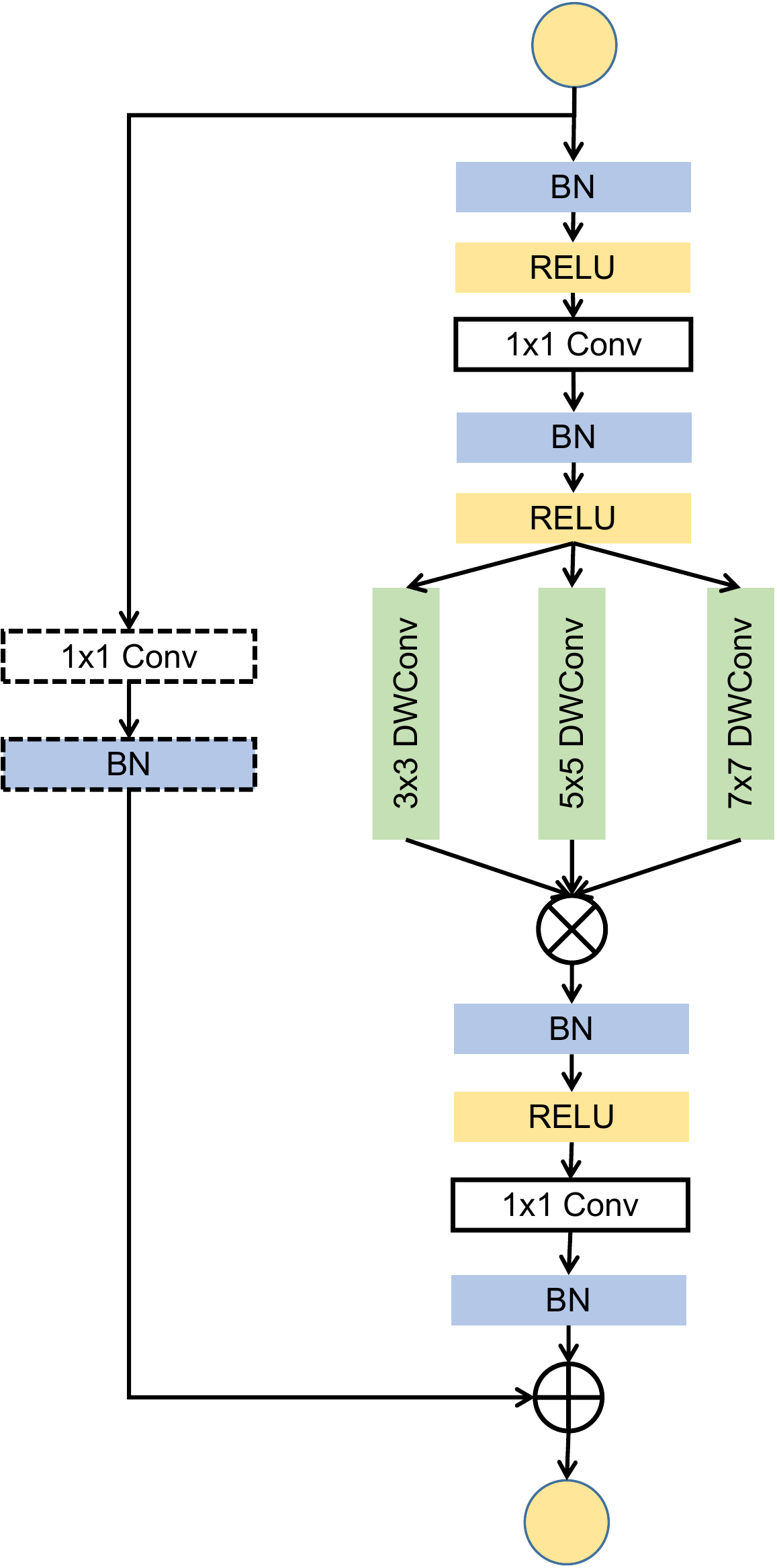}
        \label{fig:msmobile_concat}}
    \end{minipage}
    \caption{Architecture of a) Original Residual block with standard convolution; b) Residual block with Depthwise Separable Convolution; c) Residual block with Pyramid Depthwise Separable Convolution which combines features by an addition; and d) Residual block with Pyramid Depthwise Separable Convolution which combines features by a concatenation. The dash Batch Normalization and $1\times 1$ Convolution layers in the shortcut path means they do not exist in the block with stride = 1 and exist if stride = 2.}
    \label{fig:residual_architecture}
  \end{figure*}
  \subsection{Depthwise Separable Convolutions}
  Nowadays, there are many efficient neural network architectures \cite{chollet2017xception,howard2017mobilenets,sandler2018mobilenetv2,zhang2018shufflenet} use Depthwise Separable Convolutions (DWConvolution) as the key building block. The basic idea of DWConvolution is to replace a standard convolutional layer with two separate layers. The first layer uses a depthwise convolution operator. It applies a single convolutional filter per input channel to capture the spatial information in each channel. Then the second layer employs a pointwise convolution, means a $1\times 1$ convolution, to capture the cross-channel information.
  
  Suppose the input tensor $L_i$ has size $h\times w \times d_i$, the output tensor $L_j$ has size $h\times w \times d_j$. So, the standard Convolution needs to apply a convolutional kernel $K\in \mathcal{R}^{k\times k \times d_i \times d_j}$, where $k$ is the size of kernel. Therefore, it has the computation cost of $h\cdot w\cdot d_i\cdot d_j\cdot k\cdot k$.
  
  In case of DWConvolution, the depthwise convolution layer costs $h\cdot w\cdot d_i\cdot k\cdot k$ and the $1\times 1$ pointwise convolution costs $h\cdot w\cdot d_i\cdot d_j$. Hence, the total computational cost of DWConvolution is $h\cdot w\cdot d_i\cdot(k^2+d_j)$. Effectively, the computational cost of DWConvolution is smaller than the standard Convolution by a factor of $\dfrac{k^2\cdot d_j}{(k^2 + d_j)}$.
  
  \subsection{Width Multiplier: Thinner or Thicker Models}
  In real-world, there are many use cases or applications may require the model to be smaller and faster. In contrast, there will be some cases that do not care much about computation cost, the model can be fatter to achieve better results. In order to construct these smaller and less or fatter and more computationally expensive models, a very simple parameter $\alpha$, called width multiplier, is introduced.
  
  The role of the width multiplier $\alpha$ is to thin or thick a network uniformly at some layers. In CNNs, the \#channel can be changed by simply using a $1\times 1$ convolution, which is similar to bottleneck Residual module of ResNet \cite{he2016deep} or DenseNet \cite{huang2017densely}. The difference is \#channels can be reduced or increased, means $\alpha$ can be any real positive value, not just less than 1 like in bottleneck.
  
  For a given layer and width multiplier $\alpha$, the number of input channels $M$ becomes $\alpha M$ and the number of output channels $N$ becomes $\alpha N$. In case of DWConvolution with width multiplier $\alpha$, the computational cost is: $h\cdot w\cdot \alpha d_i \cdot (k^2 + d_j)$, where $\alpha\in\mathcal{R}^+$. $\alpha=1$ is the baseline networks, $\alpha < 1$ are thinner networks ($\alpha=\{0.25; 0.5; 0.75; 1\}$ in case of MobileNets), and $\alpha > 1$ are thicker networks ($\alpha=6$ in case of MobileNetsV2).
  
  Width multiplier $\alpha$ has the effect of reducing or increasing the size of network and the computational cost quadratically by roughly $\alpha^2$ in case of standard Convolution and $\alpha$ in case of DWConvolution. This parameter can be applied to any model structure to define a new smaller/bigger model with a very small change in architecture, which needs to be trained from scratch, with a reasonable accuracy, latency and size trade-off.
  
  \section{Proposed Method}
  \subsection{Pyramid Depthwise Separable Convolutions}
  The Pyramid Depthwise Separable Convolution (PydDWConvolution) uses a pyramid of kernel size $K = \{k_1, k_2, \dots, k_N\}$ for the depthwise convolution layer instead of just one kernel size. Then combines all output of this convolutions before go to the pointwise $1\times 1$ convolution. There are two ways of combining features: addition and concatenation.
  
  \paragraph{Addition.} The computation cost of $N$ depthwise convolution $K=\{k_1,k_2,\dots,k_M\}$ in case of additional combination is $h\cdot w\cdot d_i \cdot \sum_{m=0}^{M}k_m^2$. The additional operator costs $(M-1)\cdot h\cdot w \cdot d_i$. And the pointwise $1\times 1$ convolution costs $h\cdot w\cdot d_i\cdot d_j$. In summary, the computation cost of PydDWConvolution-Add is $h\cdot w\cdot d_i\cdot(M - 1 + \sum_{m=0}^{M}k_m^2 + d_j)$. So the ratio of computation cost of standard convolution and the PydDWConvolution-Add is $\dfrac{k^2\cdot d_j}{(M - 1 + \sum_{m=0}^{M}k_m^2 + d_j)}$.
  
  \paragraph{Concatenation.} The computation cost of $M$ depthwise convolution $K=\{k_1,k_2,\dots,k_M\}$ in case of concatenation combination is $h\cdot w\cdot d_i \cdot \sum_{m=0}^{N}k_m^2$. The concatenation operator costs $0$. And the pointwise $1\times 1$ convolution costs $h\cdot w\cdot N\cdot d_i\cdot d_j$. In summary, the computation cost of PydDWConvolution-Concat is $h\cdot w\cdot d_i\cdot(\sum_{m=0}^{M}k_m^2+M\cdot d_j)$. So the ratio of computation cost of standard convolution and the PydDWConvolution-Concat is $\dfrac{k^2\cdot d_j}{(\sum_{m=0}^{M}k_m^2 + M\cdot d_j)}$.
  
  As can be seen, the concatenation will increase \#parameters of model quicker than addition.
  
  \subsection{Model Architecture} \label{sec:modelarchitecture}
  This section describes the architecture of the proposed model in detail. As discussed in the previous section, the basic building block is a Residual block. The way how to apply width multiplier $\alpha$ in this block is shown in Table~\ref{tbl:residualblocktransform}. Where firstly, a $1\times 1$ convolution change \#channels by a factor $\alpha$ followed by the main convolution which can be a standard Convolution, or DWConvolution, or PydDWConvolution, with stride $=s$. Finally, another $1\times 1$ convolution is used to change \#channels to the output \#channels.
  
  The detailed architecture of different configurations of residual block are shown in Figure~\ref{fig:residual_architecture}. This paper use four configurations of Residual block. They are the Residual block with standard Convolution (Figure~\ref{fig:residual}), DWConvolution (Figure~\ref{fig:mobile}), Addition PydDWConvolution (Figure~\ref{fig:msmobile_plus}), and Concatenation PydDWConvolution (Figure~\ref{fig:msmobile_concat}).
  
  Table~\ref{tbl:structure} shows two network configurations used in this paper. They have different \#layers (Net-29 means having 29 layers and Net-56 means having 56 layers) by control number of Residual blocks.
  
  There are four kinds of networks, corresponding to four kinds of Residual block, used in experiments. They are \textit{ResNet} uses Residual block with standard $3\times 3$ Convolution; \textit{MobilenNet} uses Residual block with $3\times 3$ DWConvolution, \textit{PydMobileNet-Add} uses Residual block with Addition PydDWConvolution and \textit{PydMobileNet-Concat} uses Residual block with Concatenation PydDWConvolution. The pyramid kernel size of PydDWConvolution is \{$3\times 3; 5\times 5;$ and $7\times 7$\} 
  
  This paper also uses different value of width multiplier $\alpha$ for different configurations. $\alpha=0.5$ for ResNet; $\alpha$ with typical setting of \{0.5, 1, 1.5\} in case of MobileNet; $\alpha$ = \{0.25, 0.5, 0.75, 1\} in case of PydMobileNet-Add; and $\alpha$ = \{0.25, 0.5, 0.75\} for PydMobileNet-Concat.
  
  \section{Experiments}
  This paper evaluates own implementation of ResNet, MobileNet, and PydMobileNet on the CIFAR-10 and CIFAR-100 datasets \cite{krizhevsky2009learning} and compare with state-of-the-art architectures, especially with ResNet, ConDenseNet, and their variants. The code and models reproducing these experiments will be public later\footnote{The code and models will be public after this paper is accepted}.
  
  \subsection{Dataset}
  The two CIFAR datasets consist of RGB natural images with size $32\times 32$ pixels. The CIFAR-10 and CIFAR-100 have images drawn from 10 classes and 100 classes, respectively. These both datasets contain 50,000 images in training set and 10,000 images in testing set. This paper adopts a standard data-augmentation scheme \cite{lee2015deeply,rasmus2015semi,russakovsky2015imagenet,sermanet2013pedestrian,zagoruyko2016wide} in which the training images are random horizontal mirroring and zero-padded with 4 pixels on each side, randomly cropped to produce the original $32\times 32$ pixels size.
  \begin{table*}[!t]
    \def\arraystretch{1}%
    \setlength\tabcolsep{10pt}
    \caption{Error rates (\%) on CIFAR-10 and CIFAR-100 datasets of own implemented models. * indicates models obtained from GluonCV toolkit of MXNet\protect\footnotemark. Results of PydMobileNet that outperform ResNet and MobileNet at the same \#layers are \textbf{bold} and the overal best results are \textbf{\color{blue} blue}. The model names also contain the \#layers and width multiplier $\alpha$, in turn. The proposed PydMobileNets achieve lower error rates while using fewer parameters than ResNets and MobileNets.}
    \label{tbl:ownexperiments}
    \centering
    \begin{tabular}{l||c|c|c|c|c}
      \hline
      Model & Depth & \#Params & FLOPs & CIFAR-10 & CIFAR-100\\\hline
      \hline
      ResNet-20* & 20 & 0.278M & 87M & 7.3 & -\\
      \hline
      {ResNet-29-0.5} & 29 & 0.221M & 29M & 6.97 & 19.62\\\hline
      {MobileNet-29-0.5} & 29 & 0.079M & 12M & 8.63 & 22.59 \\
      {MobileNet-29-1} & 29 & 0.142M & 22M & 7.09 & 19.40\\
      {MobileNet-29-1.5} & 29 & 0.206M & 32M & 6.56 & 18.09\\\hline
      {PydMobileNet-Add-29-0.25} & 29 & 0.060M & 10M & 9.43 & 21.96\\
      {PydMobileNet-Add-29-0.5} & 29 & 0.104M & 18M & 7.29 & 20.26\\
      {PydMobileNet-Add-29-0.75} & 29 & 0.148M & 26M & \textbf{6.52} & \textbf{17.95}\\
      {PydMobileNet-Add-29-1} & 29 & 0.193M & 34M & \textbf{6.00} & \textbf{17.54}\\\hline
      {PydMobileNet-Concat-29-0.25} & 29 & 0.092M & 14M & 7.33 & 21.04\\
      {PydMobileNet-Concat-29-0.5} & 29 & 0.170M & 27M & \textbf{5.71} & \textbf{17.27}\\
      {PydMobileNet-Concat-29-0.75} & 29 & 0.247M & 39M & \textbf{\color{blue} 5.68} & \textbf{\color{blue} 16.28}\\\hline
      \hline
      ResNet-56* & 56 & 0.861M & 277M & 5.4 & -\\\hline
      {ResNet-56-0.5} & 56 & 0.435M & 60M & 5.76 & 17.60\\\hline
      {MobileNet-56-0.5} & 56 & 0.151M & 23M & 6.75 & 18.56\\
      {MobileNet-56-1} & 56 & 0.283M & 43M & 6.02 & 17.15\\
      {MobileNet-56-1.5} & 56 & 0.416M & 63M & 5.29 & 16.58\\\hline
      {PydMobileNet-Add-56-0.25} & 56 & 0.109M & 19M & 7.38 & 20.41\\
      {PydMobileNet-Add-56-0.5} & 56 & 0.200M & 36M & 6.19 & 17.36\\
      {PydMobileNet-Add-56-0.75} & 56 & 0.292M & 52M & 5.55 & \textbf{16.58}\\
      {PydMobileNet-Add-56-1} & 56 & 0.382M & 69M & \textbf{4.98} & \textbf{16.23}\\\hline
      {PydMobileNet-Concat-56-0.25} & 56 & 0.175M & 28M & 6.23 & 17.85\\
      {PydMobileNet-Concat-56-0.5} & 56 & 0.332M & 53M & \textbf{5.24} & \textbf{15.67}\\
      {PydMobileNet-Concat-56-0.75} & 56 & 0.489M & 79M & \textbf{\color{blue} 4.72} & \textbf{\color{blue} 14.60}\\\hline
      \multicolumn{5}{l}{\footnotesize{$^2$ https://gluon-cv.mxnet.io/model\_zoo/classification.html\#cifar10}}
    \end{tabular}
  \end{table*}
  \begin{figure}[!t]
    \centering
    \includegraphics[width=\linewidth]{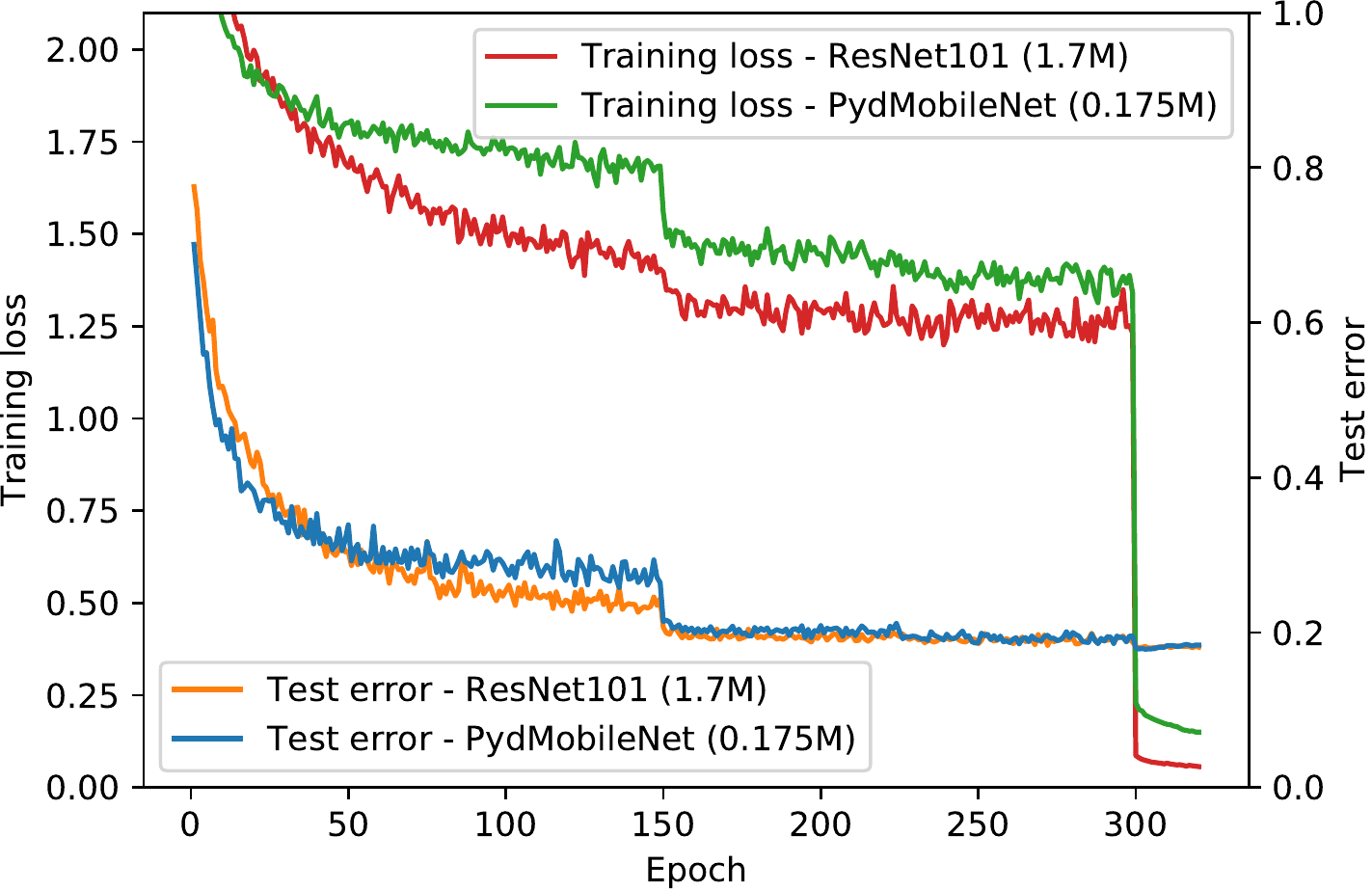}
    \caption{Training and testing curves of the 110-layer ResNet \cite{he2016identity} with more than 1.7M parameters and a 56-layer PydMobileNet-0.25-Concat with only 0.175M parameters.}
    \label{fig:loss_acc}
  \end{figure}
  \subsection{Implementation Details}
  This paper implements all networks on Gluon module of MXNet open source deep learning framework \cite{chen2015mxnet}. The training procedure follows the schema proposed in \cite{zhang2017mixup}. All models are trained using back-propagation \cite{lecun1989backpropagation} by Stochastic Gradient Descent \cite{robbins1985stochastic} with Nesterov momentum \cite{nesterov1983method} (NAG) optimizer implemented by MXNet for 320 epochs. The initial learning rate is set to 0.1 and is reduced 10 times at 150 and 225 epochs, respectively. The parameters are initialized by Xavier's initializer \cite{glorot2010understanding}. The other settings are: weight decay of 0.0001, momentum of 0.9, and batch size of 128.
  
  \subsection{Performance Evaluation}
  This paper uses the top-1 error rate for evaluating proposed network architecture. The ResNets, MobileNets, and PydMobileNets are trained based on the configurations already mentioned in the previous section.
  \begin{table*}[!t]
    \def\arraystretch{1.1}%
    \setlength\tabcolsep{10pt}
    \caption{Error rates (\%) on CIFAR-10 and CIFAR-100 datasets. Results that outperform all competing methods are \textbf{bold} and the overal best results are \textbf{\color{blue} blue}. FLOPs information is obtained from \cite{huang2018condensenet}. * indicates models obtained from GluonCV toolkit of MXNet\protect\footnotemark. $k$ in DenseNet \cite{huang2017densely} denotes network's growth rate. \textit{Italic} names indicate models run by ourselves. Where \textit{ResNet}, \textit{MobileNet}, \textit{PydMobileNet-Add}, and \textit{PydMobileNet-Concat} use Residual block with standard Convolution, DWConvolution, Addition  Pyramid DWConvolution, and Concatenation DWConvolution, respectively. The model names also contain the \#layers and width multiplier $\alpha$, in turn. The proposed PydMobileNets achieve similar or even lower error rates while using much fewer parameters than other networks.}
    \label{tbl:comparedexperiments}
    \centering
    \begin{tabular}{l||c|c|c|c|c}
      \hline
      Model & Depth & \#Params & FLOPs & CIFAR-10 & CIFAR-100\\\hline
      \hline
      Network in Network \cite{lin2013network} & - & - & - & 8.81 & - \\
      All-CNN \cite{springenberg2015striving} & - & - & - & 7.25 & 33.71 \\
      Deeply Supervised Net \cite{lee2015deeply} & - & - & - & 7.97 & 34.57 \\
      Highway Network \cite{srivastava2015training} & - & - & - & 7.72 & 32.39 \\\hline
      FractalNet \cite{larsson2016fractalnet} & 21 & 38.6M & - & 5.22 & 23.30 \\
      with Dropout/Drop-path & 21 & 38.6M & - & 4.60 & 23.73 \\\hline
      ResNet \cite{he2016deep} & 110 & 1.7M & - & 6.61 & - \\\hline
      ResNet (reported by \cite{huang2016deep}) & 110 & 1.7M & - & 6.41 & 27.22 \\\hline
      ResNet with Stochastic Depth \cite{huang2016deep} & 110 & 1.7M & - & 5.23 & 24.58 \\
      {} & 1202 & 19.4M & 2,840M & 4.91 & - \\\hline
      Wide ResNet \cite{zagoruyko2016wide} & 16 & 11.0M & - & 4.81 & 22.07 \\
      {} & 28 & 36.5M & 5,248M & 4.17 & 20.50 \\\hline
      ResNet (pre-activation) \cite{he2016identity} & 164 & 1.7M & - & 5.46 & 24.33 \\
      {} & 1001 & 16.1M & 2,357M & 4.62 & 22.71 \\\hline
      ResNeXt-29 \cite{xie2017aggregated} & 29 & 68.1M & 10,704M & 3.58 & 17.31\\\hline
      NASNet-A \cite{zoph2017learning} & - & 3.3M & - & 3.41 & -\\\hline
      DenseNet-BC ($k = 12$) \cite{huang2017densely} & 100 & 0.8M & - & 4.51 & 22.27 \\
      DenseNet-BC ($k = 24$) & 250 & 15.3M & - & 3.62 & 17.60\\
      DenseNet-BC ($k = 40$) & 190 & 25.6M & 9,388M & \textbf{\color{blue} 3.46} & 17.18\\\hline
      CondenseNet$^\text{light}$-94 \cite{huang2018condensenet} & 94 & 0.33M & 122M & 5.00 & 24.08\\
      CondenseNet-86 & 86 & 0.52M & 65M & 5.00 & 23.64\\
      CondenseNet$^\text{light}$-160 & 160 & 3.1M & 1,084M & \textbf{\color{blue} 3.46} & 17.55\\
      CondenseNet-182 & 182 & 4.2M & 513M & 3.76 & 18.47\\\hline
      ResNet-110* & 110 & 1.736M & 562M & 4.50 & 18.00\\\hline
      \hline
      {PydMobileNet-Add-29-0.25} & 29 & 0.060M & 10M & 9.43 & 21.96\\
      {PydMobileNet-Add-29-0.5} & 29 & 0.104M & 18M & 7.29 & 20.26\\
      {PydMobileNet-Add-29-0.75} & 29 & 0.148M & 26M & 6.52 & 17.95\\
      {PydMobileNet-Add-29-1} & 29 & 0.193M & 34M & 6.00 & 17.54\\\hline
      {PydMobileNet-Concat-29-0.25} & 29 & 0.092M & 14M & 7.33 & 21.04\\
      {PydMobileNet-Concat-29-0.5} & 29 & 0.170M & 27M & 5.71 & 17.27\\
      {PydMobileNet-Concat-29-0.75} & 29 & 0.247M & 39M & 5.68 & \textbf{16.28}\\\hline
      \hline
      {PydMobileNet-Add-56-0.25} & 56 & 0.109M & 19M & 7.38 & 20.41\\
      {PydMobileNet-Add-56-0.5} & 56 & 0.200M & 36M & 6.19 & 17.36\\
      {PydMobileNet-Add-56-0.75} & 56 & 0.292M & 52M & 5.55 & \textbf{16.58}\\
      {PydMobileNet-Add-56-1} & 56 & 0.382M & 69M & 4.98 & \textbf{16.23}\\\hline
      {PydMobileNet-Concat-56-0.25} & 56 & 0.175M & 28M & 6.23 & 17.85\\
      {PydMobileNet-Concat-56-0.5} & 56 & 0.332M & 53M & 5.24 & \textbf{15.67}\\
      {PydMobileNet-Concat-56-0.75} & 56 & 0.489M & 79M & 4.72 & \textbf{\color{blue} 14.60}\\\hline
      \multicolumn{5}{l}{\footnotesize{$^3$ https://gluon-cv.mxnet.io/model\_zoo/classification.html\#cifar10}}
    \end{tabular}
  \end{table*}
  \begin{table*}[!t]
    \def\arraystretch{1}%
    \setlength\tabcolsep{10pt}
    \caption{Inference speed of own implemented models when running on CPU and GPU with batch size of 128.}
    \label{tbl:actualspeed}
    \centering
    \begin{tabular}{l||c|c|c|c}
      \hline
      Model & \#Params & FLOPs & GPU time (\textit{ms}) & CPU time (\textit{ms})\\\hline
      {ResNet-56-0.5} & 0.435M & 60M &  75 & 202\\\hline
      {MobileNet-56-0.5} & 0.151M & 23M & 76 & 177\\
      {MobileNet-56-1} & 0.283M & 43M & 107 & 261\\
      {MobileNet-56-1.5} & 0.416M & 63M & 162 & 353\\\hline
      {PydMobileNet-Add-56-0.25} & 0.109M & 19M & 84 & 208\\
      {PydMobileNet-Add-56-0.5} & 0.200M & 36M & 127 & 252\\
      {PydMobileNet-Add-56-0.75} & 0.292M & 52M & 175 & 308\\
      {PydMobileNet-Add-56-1} & 0.382M & 69M & 221 & 378\\\hline
      {PydMobileNet-Concat-56-0.25} & 0.175M & 28M & 92 & 207\\
      {PydMobileNet-Concat-56-0.5} & 0.332M & 53M & 150 & 306\\
      {PydMobileNet-Concat-56-0.75} & 0.489M & 79M & 207 & 405\\\hline
    \end{tabular}
  \end{table*}
  \subsubsection{Comparison between Different Residual Block Structure}
  Table~\ref{tbl:ownexperiments} shows the comparison between own implemented models mentioned in Section~\ref{sec:modelarchitecture}. The comparison is fair because all models have similar structure, they are just different together in Residual block structure. To highlight general trends, all results of PydMobileNets that outperform the ResNets and MobileNets are in \textbf{bold} and the overall best results are in \textbf{\color{blue} blue}.
  
  As can be seen, the PydMobileNets outperform other models in the same \#layers. A very clear trend is if width multiplier $\alpha$ increases, the \#parameters increases and the error rate decreases in both MobileNets and PydMobileNets. The PydMobileNets are slightly better than MobileNets with similar \#parameters. This situation is natural because they can capture more spatial information.
  
  In comparison between addition and concatenation when combining features, their performances are comparable in term of similar \#parameters. The concatenation increase \#parameters quicker.
  
  So, the capacity of models can be easily controlled by adjusting \#layers, value of width multiplier $\alpha$, and the way of combination. This helps PydMobileNet can be more flexible in fine-tuning the trade-off between accuracy, latency, and model size.
  
  The two ResNets obtained from GluonCV toolkit of MXNet use the original Residual block with two $3\times 3$ standard Convolution \cite{he2016identity}. One interesting thing here is the ResNet uses Residual block with bottleneck designed by this paper can achieve similar performance with much more compact models. It is an additional evidence for the fact that the bottleneck modules can be a simple way to compress model.
  
  \subsubsection{Comparison with Other Models}
  Table~\ref{tbl:comparedexperiments} shows the comparison between own implemented models and other models. Similarly to previous section, all results of PydMobileNets that outperform all existing models are in \textbf{bold} and the overall best results are in \textbf{\color{blue} blue}, to highlight general trends.
  
  As can be seen, the PydMobileNets outperform other state-of-the-art models in CIFAR-100 dataset and achieve similar error rate in CIFAR-10 dataset while having much fewer \#parameters. Figure~\ref{fig:loss_acc} shows the training loss and test errors of 110-layer ResNets and 56-layer PydMobileNet-0.25-Concat on CIFAR-10 datasets. The 110-layer deep ResNet converges to a lower training loss value but a similar test error.
  
  \subsubsection{Actual Inference Time Evaluation}
  Finally, this paper evaluates the actual inference speed of own implemented models: ResNet, MobileNet, and PydMobileNet on a computer with Intel Core i7-4770 3.40-GHz CPU, NVIDIA 750Ti GPU, and 8-GB RAM. The evaluation is done for networks have 56 layers with batch size 128 to show the difference more straightforwardly. It reports running speed on both CPU and GPU. The results are shown in Table~\ref{tbl:actualspeed}.
  
  As can be seen, the MobileNets are slow than ResNet in both CPU and GPU although the FLOPs is smaller. Because the Depthwise Separable Convolution is not (yet) efficiently implemented in MXNet.
  
  In comparison between PydMobileNet's variants, the concatenation looks more efficient than addition. Maybe the reason is from the worse computation/memory access ratio in compared with concatenation. The speed of PydMobileNet-Concats are similar with MobileNets in term of similar \#parameters. Therefore, this paper suggests concatenation should be used in real applications.
  
  \section{Conclusion}
  This paper introduced an improved version of MobileNet, called PydMobileNet, which use pyramid kernel size in DWConvolution instead of just a DWConvolution.
  This helps network can capture more spatial information. Additionally, by adjusting the width multiplier and the way of combining features, the capacity of the network can be easily controlled, which helps PydMobilnet can be used in many use cases.
  
  The experiments showed that the PydMobileNets can achieve similar or even lower error rate with much fewer \#parameters in comparing to MobileNets as well as other state-of-the-art methods.
  
  In the future, it is necessary to evaluate proposed architecture with more experiments on the ImageNet dataset \cite{russakovsky2015imagenet}. Additionally, the atrous Convolution should be considered because it is an efficient way to capture difference spatial information without increasing computational cost much.
  {\small
    \bibliographystyle{ieee}
    \bibliography{1Bibliography}

\begin{thebibliography}{10}\itemsep=-1pt

\bibitem{alvarez2016learning}
J.~M. Alvarez and M.~Salzmann.
\newblock Learning the number of neurons in deep networks.
\newblock In {\em Proceedings of the Advances in Neural Information Processing
  Systems}, pages 2270--2278, 2016.

\bibitem{bolukbasi2017adaptive}
T.~Bolukbasi, J.~Wang, O.~Dekel, and V.~Saligrama.
\newblock Adaptive neural networks for fast test-time prediction.
\newblock 2017.

\bibitem{buciluǎ2006model}
C.~Buciluǎ, R.~Caruana, and A.~Niculescu-Mizil.
\newblock Model compression.
\newblock In {\em Proceedings of the ACM SIGKDD International Conference on
  Knowledge Discovery and Data Mining}, pages 535--541. ACM, 2006.

\bibitem{chen2015mxnet}
T.~Chen, M.~Li, Y.~Li, M.~Lin, N.~Wang, M.~Wang, T.~Xiao, B.~Xu, C.~Zhang, and
  Z.~Zhang.
\newblock Mxnet: A flexible and efficient machine learning library for
  heterogeneous distributed systems.
\newblock {\em Neural Information Processing Systems, Workshop on Machine
  Learning Systems}, 2015.

\bibitem{chen2015compressing}
W.~Chen, J.~Wilson, S.~Tyree, K.~Weinberger, and Y.~Chen.
\newblock Compressing neural networks with the hashing trick.
\newblock In {\em Proceedings of the International Conference on Machine
  Learning}, pages 2285--2294, 2015.

\bibitem{chollet2017xception}
F.~Chollet.
\newblock Xception: Deep learning with depthwise separable convolutions.
\newblock In {\em Proceedings of the IEEE Conference on Computer Vision and
  Pattern Recognition}, pages 1800--1807. IEEE, 2017.

\bibitem{figurnov2017spatially}
M.~Figurnov, M.~D. Collins, Y.~Zhu, L.~Zhang, J.~Huang, D.~P. Vetrov, and
  R.~Salakhutdinov.
\newblock Spatially adaptive computation time for residual networks.
\newblock In {\em Proceedings of the IEEE Conference on Computer Vision and
  Pattern Recognition}, 2017.

\bibitem{girshick2014rich}
R.~Girshick, J.~Donahue, T.~Darrell, and J.~Malik.
\newblock Rich feature hierarchies for accurate object detection and semantic
  segmentation.
\newblock In {\em Proceedings of the IEEE Conference on Computer Vision and
  Pattern Recognition}, pages 580--587, 2014.

\bibitem{glorot2010understanding}
X.~Glorot and Y.~Bengio.
\newblock Understanding the difficulty of training deep feedforward neural
  networks.
\newblock In {\em Proceedings of the International Conference on Artificial
  Intelligence and Statistics}, pages 249--256, 2010.

\bibitem{han2015deep}
S.~Han, H.~Mao, and W.~J. Dally.
\newblock Deep compression: Compressing deep neural networks with pruning,
  trained quantization and huffman coding.
\newblock {\em Proceedings of the International Conference on Machine
  Learning}, 2016.

\bibitem{han2015learning}
S.~Han, J.~Pool, J.~Tran, and W.~Dally.
\newblock Learning both weights and connections for efficient neural network.
\newblock In {\em Proceedings of the Advances in Neural Information Processing
  Systems}, pages 1135--1143, 2015.

\bibitem{hassibi1993optimal}
B.~Hassibi, D.~G. Stork, and G.~J. Wolff.
\newblock Optimal brain surgeon and general network pruning.
\newblock In {\em Proceedings of the IEEE International Conference on Neural
  Networks}, pages 293--299. IEEE, 1993.

\bibitem{he2016deep}
K.~He, X.~Zhang, S.~Ren, and J.~Sun.
\newblock Deep residual learning for image recognition.
\newblock In {\em Proceedings of the IEEE Conference on Computer Vision and
  Pattern Recognition}, pages 770--778, 2016.

\bibitem{he2016identity}
K.~He, X.~Zhang, S.~Ren, and J.~Sun.
\newblock Identity mappings in deep residual networks.
\newblock In {\em Proceedings of the European Conference on Computer Vision},
  pages 630--645. Springer, 2016.

\bibitem{he2017channel}
Y.~He, X.~Zhang, and J.~Sun.
\newblock Channel pruning for accelerating very deep neural networks.
\newblock In {\em Proceedings of the IEEE International Conference on Computer
  Vision}, volume~2, 2017.

\bibitem{hinton2014distilling}
G.~Hinton, O.~Vinyals, and J.~Dean.
\newblock Distilling the knowledge in a neural network.
\newblock {\em Proceedings of the Advances in Neural Information Processing
  Systems Deep Learning Workshop}, 2014.

\bibitem{howard2017mobilenets}
A.~G. Howard, M.~Zhu, B.~Chen, D.~Kalenichenko, W.~Wang, T.~Weyand,
  M.~Andreetto, and H.~Adam.
\newblock Mobilenets: Efficient convolutional neural networks for mobile vision
  applications, 2017.

\bibitem{huang2017multi}
G.~Huang, D.~Chen, T.~Li, F.~Wu, L.~van~der Maaten, and K.~Q. Weinberger.
\newblock Multi-scale dense networks for resource efficient image
  classification.
\newblock 2018.

\bibitem{huang2018condensenet}
G.~Huang, S.~Liu, L.~van~der Maaten, and K.~Q. Weinberger.
\newblock Condensenet: An efficient densenet using learned group convolutions.
\newblock {\em Proceedings of the IEEE Conference on Computer Vision and
  Pattern Recognition}, 2018.

\bibitem{huang2017densely}
G.~Huang, Z.~Liu, L.~Van Der~Maaten, and K.~Q. Weinberger.
\newblock Densely connected convolutional networks.
\newblock In {\em Proceedings of the IEEE Conference on Computer Vision and
  Pattern Recognition}, 2017.

\bibitem{huang2016deep}
G.~Huang, Y.~Sun, Z.~Liu, D.~Sedra, and K.~Q. Weinberger.
\newblock Deep networks with stochastic depth.
\newblock In {\em Proceedings of the European Conference on Computer Vision},
  pages 646--661. Springer, 2016.

\bibitem{hubara2016binarized}
I.~Hubara, M.~Courbariaux, D.~Soudry, R.~El-Yaniv, and Y.~Bengio.
\newblock Binarized neural networks.
\newblock In {\em Proceedings of the Advances in neural information processing
  systems}, pages 4107--4115, 2016.

\bibitem{iandola2016squeezenet}
F.~N. Iandola, S.~Han, M.~W. Moskewicz, K.~Ashraf, W.~J. Dally, and K.~Keutzer.
\newblock Squeezenet: Alexnet-level accuracy with 50x fewer parameters and< 0.5
  mb model size, 2016.

\bibitem{jaderberg2014speeding}
M.~Jaderberg, A.~Vedaldi, and A.~Zisserman.
\newblock Speeding up convolutional neural networks with low rank expansions.
\newblock In {\em Proceedings of the British Machine Vision Conference. BMVA
  Press}, 2014.

\bibitem{krizhevsky2009learning}
A.~Krizhevsky and G.~Hinton.
\newblock Learning multiple layers of features from tiny images.
\newblock Technical report, University of Toronto, 2009.

\bibitem{larsson2016fractalnet}
G.~Larsson, M.~Maire, and G.~Shakhnarovich.
\newblock Fractalnet: Ultra-deep neural networks without residuals, 2016.

\bibitem{lecun1989backpropagation}
Y.~LeCun, B.~Boser, J.~S. Denker, D.~Henderson, R.~E. Howard, W.~Hubbard, and
  L.~D. Jackel.
\newblock Backpropagation applied to handwritten zip code recognition.
\newblock {\em Neural Computation}, 1(4):541--551, 1989.

\bibitem{lecun1990optimal}
Y.~LeCun, J.~S. Denker, and S.~A. Solla.
\newblock Optimal brain damage.
\newblock In {\em Proceedings of the Advances in neural information processing
  systems}, pages 598--605, 1990.

\bibitem{lee2015deeply}
C.-Y. Lee, S.~Xie, P.~Gallagher, Z.~Zhang, and Z.~Tu.
\newblock Deeply-supervised nets.
\newblock In {\em Proceedings of the International Conference on Artificial
  Intelligence and Statistics}, pages 562--570, 2015.

\bibitem{lin2013network}
M.~Lin, Q.~Chen, and S.~Yan.
\newblock Network in network.
\newblock In {\em Proceedings of the International Conference on Learning
  Representations}, 2014.

\bibitem{mao2017exploring}
H.~Mao, S.~Han, J.~Pool, W.~Li, X.~Liu, Y.~Wang, and W.~J. Dally.
\newblock Exploring the regularity of sparse structure in convolutional neural
  networks, 2017.

\bibitem{nesterov1983method}
Y.~E. Nesterov.
\newblock A method for solving the convex programming problem with convergence
  rate o (1/k\^{} 2).
\newblock In {\em Dokl. Akad. Nauk SSSR}, volume 269, pages 543--547, 1983.

\bibitem{radosavovic2018data}
I.~Radosavovic, P.~Doll{\'a}r, R.~Girshick, G.~Gkioxari, and K.~He.
\newblock Data distillation: Towards omni-supervised learning.
\newblock In {\em Proceedings of the IEEE Conference on Computer Vision and
  Pattern Recognition}, pages 4119--4128, 2018.

\bibitem{rasmus2015semi}
A.~Rasmus, M.~Berglund, M.~Honkala, H.~Valpola, and T.~Raiko.
\newblock Semi-supervised learning with ladder networks.
\newblock In {\em Proceedings of the Advances in Neural Information Processing
  Systems}, pages 3546--3554, 2015.

\bibitem{rastegari2016xnor}
M.~Rastegari, V.~Ordonez, J.~Redmon, and A.~Farhadi.
\newblock Xnor-net: Imagenet classification using binary convolutional neural
  networks.
\newblock In {\em Proceedings of the European Conference on Computer Vision},
  pages 525--542. Springer, 2016.

\bibitem{robbins1985stochastic}
H.~Robbins and S.~Monro.
\newblock A stochastic approximation method.
\newblock In {\em Herbert Robbins Selected Papers}, pages 102--109. Springer,
  1985.

\bibitem{russakovsky2015imagenet}
O.~Russakovsky, J.~Deng, H.~Su, J.~Krause, S.~Satheesh, S.~Ma, Z.~Huang,
  A.~Karpathy, A.~Khosla, M.~Bernstein, et~al.
\newblock Imagenet large scale visual recognition challenge.
\newblock {\em International Journal of Computer Vision}, 115(3):211--252,
  2015.

\bibitem{sandler2018mobilenetv2}
M.~Sandler, A.~Howard, M.~Zhu, A.~Zhmoginov, and L.-C. Chen.
\newblock Mobilenetv2: Inverted residuals and linear bottlenecks.
\newblock In {\em Proceedings of the IEEE Conference on Computer Vision and
  Pattern Recognition}, pages 4510--4520, 2018.

\bibitem{see2016compression}
A.~See, M.-T. Luong, and C.~D. Manning.
\newblock Compression of neural machine translation models via pruning.
\newblock In {\em Proceedings of The 20th SIGNLL Conference on Computational
  Natural Language Learning}, pages 291--301, 2016.

\bibitem{sermanet2013pedestrian}
P.~Sermanet, K.~Kavukcuoglu, S.~Chintala, and Y.~LeCun.
\newblock Pedestrian detection with unsupervised multi-stage feature learning.
\newblock In {\em Proceedings of the IEEE Conference on Computer Vision and
  Pattern Recognition}, pages 3626--3633, 2013.

\bibitem{springenberg2015striving}
J.~Springenberg, A.~Dosovitskiy, T.~Brox, and M.~Riedmiller.
\newblock Striving for simplicity: The all convolutional net.
\newblock In {\em Proceedings of the International Conference on Learning
  Representations (workshop track)}, 2015.

\bibitem{srivastava2015training}
R.~K. Srivastava, K.~Greff, and J.~Schmidhuber.
\newblock Training very deep networks.
\newblock In {\em Proceedings of the Advances in neural information processing
  systems}, pages 2377--2385, 2015.

\bibitem{szegedy2015going}
C.~Szegedy, W.~Liu, Y.~Jia, P.~Sermanet, S.~Reed, D.~Anguelov, D.~Erhan,
  V.~Vanhoucke, and A.~Rabinovich.
\newblock Going deeper with convolutions.
\newblock In {\em Proceedings of the IEEE Conference on Computer Vision and
  Pattern Recognition}, pages 1--9, 2015.

\bibitem{szegedy2016rethinking}
C.~Szegedy, V.~Vanhoucke, S.~Ioffe, J.~Shlens, and Z.~Wojna.
\newblock Rethinking the inception architecture for computer vision.
\newblock In {\em Proceedings of the IEEE Conference on Computer Vision and
  Pattern Recognition}, pages 2818--2826, 2016.

\bibitem{wu2016quantized}
J.~Wu, C.~Leng, Y.~Wang, Q.~Hu, and J.~Cheng.
\newblock Quantized convolutional neural networks for mobile devices.
\newblock In {\em Proceedings of the IEEE Conference on Computer Vision and
  Pattern Recognition}, pages 4820--4828, 2016.

\bibitem{xie2017aggregated}
S.~Xie, R.~Girshick, P.~Doll{\'a}r, Z.~Tu, and K.~He.
\newblock Aggregated residual transformations for deep neural networks.
\newblock In {\em Proceedings of the IEEE Conference on Computer Vision and
  Pattern Recognition}, pages 5987--5995. IEEE, 2017.

\bibitem{zagoruyko2016wide}
S.~Zagoruyko and N.~Komodakis.
\newblock Wide residual networks, 2016.

\bibitem{zhang2017mixup}
H.~Zhang, M.~Cisse, Y.~N. Dauphin, and D.~Lopez-Paz.
\newblock mixup: Beyond empirical risk minimization, 2017.

\bibitem{zhang2017interleaved}
T.~Zhang, G.-J. Qi, B.~Xiao, and J.~Wang.
\newblock Interleaved group convolutions for deep neural networks.
\newblock In {\em Proceedings of the IEEE International Conference on Computer
  Vision}, 2017.

\bibitem{zhang2018shufflenet}
X.~Zhang, X.~Zhou, M.~Lin, and J.~Sun.
\newblock Shufflenet: An extremely efficient convolutional neural network for
  mobile devices.
\newblock In {\em Proceedings of the IEEE Conference on Computer Vision and
  Pattern Recognition}, June 2018.

\bibitem{zoph2017learning}
B.~Zoph, V.~Vasudevan, J.~Shlens, and Q.~V. Le.
\newblock Learning transferable architectures for scalable image recognition.
\newblock {\em Proceedings of the IEEE Conference on Computer Vision and
  Pattern Recognition}, 2018.

\end{thebibliography}
  }

\end{document}